\documentclass[10pt,twocolumn,letterpaper]{article}

\usepackage[pagenumbers]{cvpr} 

\usepackage{graphicx}
\usepackage{amsmath}
\usepackage{amssymb}
\usepackage{booktabs}

\usepackage[pagebackref,breaklinks,colorlinks]{hyperref}

\usepackage[capitalize]{cleveref}
\crefname{section}{Sec.}{Secs.}
\Crefname{section}{Section}{Sections}
\Crefname{table}{Table}{Tables}
\crefname{table}{Tab.}{Tabs.}

\def\cvprPaperID{1884} 
\def\confName{CVPR}
\def\confYear{2023}

\begin{document}

\title{Multi-manifold Attention for Vision Transformers}

\author{Dimitrios Konstantinidis, Ilias Papastratis, Kosmas Dimitropoulos, Petros Daras\\
Information Technologies Institute, Centre for Research and Technology Hellas\\
6th km Charilaou-Thermi Rd, 57001 Thermi, Thessaloniki, Greece\\
{\tt\small \{dikonsta, papastrat, dimitrop, daras\}@iti.gr}
}
\maketitle

\begin{abstract}
Vision Transformers are very popular nowadays due to their state-of-the-art performance in several computer vision tasks, such as image classification and action recognition. Although their performance has been greatly enhanced through highly descriptive patch embeddings and hierarchical structures, there is still limited research on utilizing additional data representations so as to refine the self-attention map of a Transformer. To address this problem, a novel attention mechanism, called multi-manifold multi-head attention, is proposed in this work to substitute the vanilla self-attention of a Transformer. The proposed mechanism models the input space in three distinct manifolds, namely Euclidean, Symmetric Positive Definite and Grassmann, thus leveraging different statistical and geometrical properties of the input for the computation of a highly descriptive attention map. In this way, the proposed attention mechanism can guide a Vision Transformer to become more attentive towards important appearance, color and texture features of an image, leading to improved classification and segmentation results, as shown by the experimental results on well-known datasets.
\end{abstract}

\section{Introduction}
\label{sec:intro}

Transformer networks have been met with great interest from the research community when they were originally proposed for natural language processing in the pioneering work of \cite{Author28}. A Transformer is a network architecture that relies on self-attention, which is an attention mechanism that relates different positions of a single sequence in order to compute a new representation of the sequence. Transformers are designed to handle sequential input data, however, unlike Recurrent Neural Networks (RNNs), they process the data in parallel since they receive the entire sequence as input. In this way, Transformers can extract both short- and long-term dependencies between input and output, while simultaneously achieving increased parallelization and training speed with respect to RNNs.

Recently, Vision Transformers (ViTs) \cite{Author3} have been introduced for computer vision tasks, leading to state-of-the-art performance in well-known benchmark datasets due to their ability to model short- and long-range relationships between different image areas. However, ViTs lack the local inductive biases of Convolutional Neural Networks (CNNs) and do not efficiently model local information \cite{Author5}. To overcome such issues, recent works aim to increase local structure modelling by introducing convolutions to the ViTs \cite{Author1,Author5,Author6}, redesign the patch tokenization process and introduce local attention mechanisms \cite{Author8,Author9,Author48,Author50} or adopt hierarchical structures similar to CNNs \cite{Author8,Author10,Author11}.

\begin{figure}[t]
  \centering
  \includegraphics[width=\linewidth]{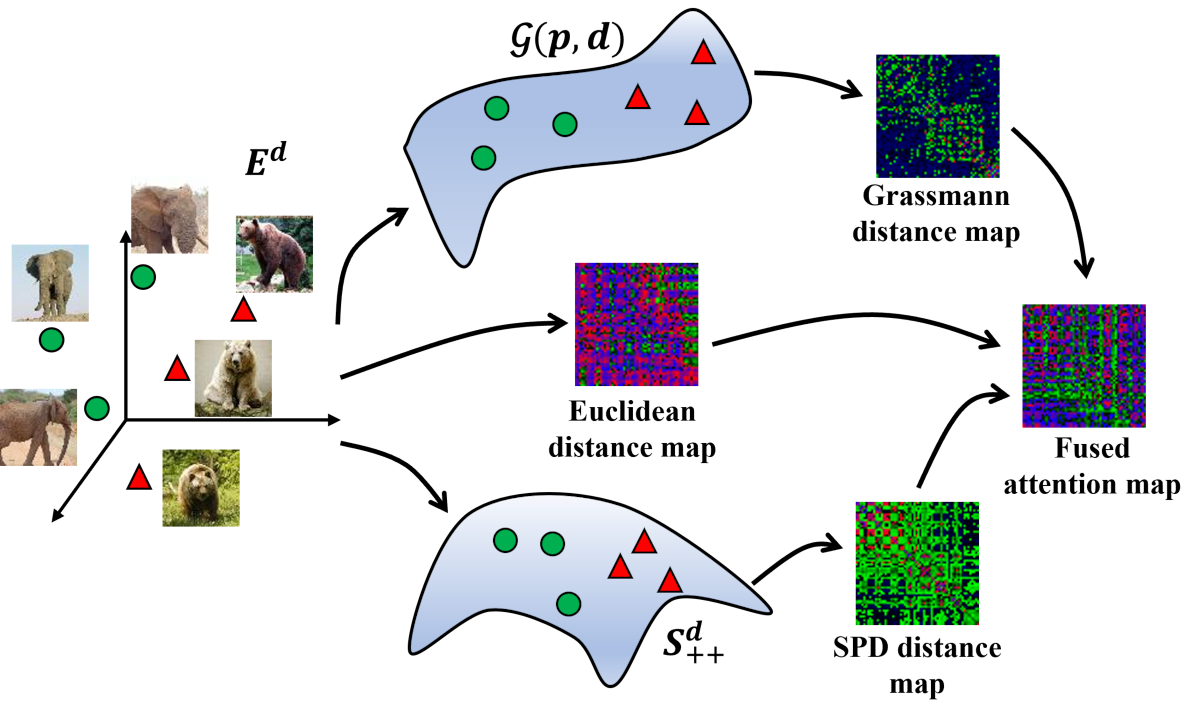}
  \caption{With the modelling of input in the Euclidean $E^d$, Grassmann $\mathcal{G}(p,d)$ and SPD $\mathcal{S}_{++}^d$ manifolds and the computation of the respective distance maps, the proposed MMA produces a fused attention map with high discriminative power.}
  \label{fig:motivation}
\end{figure}

However, most ViTs operate only in the Euclidean space of pixel intensity values, overlooking the fact that data representations in other manifolds can be beneficial to their performance. Additionally, the attention maps tend to be similar in the deeper layers of ViTs and their representation stops improving \cite{Author30}, thus a need for more discriminative attention maps is imperative. To this end, this work proposes multi-manifold multi-head attention (MMA) that can be introduced to any ViT to improve the discriminative power of its attention map. The main advantage of MMA over the vanilla self-attention is the use of three different manifolds, namely Euclidean, Symmetric Positive Definite (SPD) and Grassmann to model the input. Computing feature representations and distances in manifolds with different statistical and geometrical properties enables a better modelling of the input through the computation of an attention map that can better enhance the important features of the input for a given computer vision task, as shown in Fig. \ref{fig:motivation}. More specifically, the contributions of this work are the following:
\begin{itemize}
\item A novel self-attention mechanism, called MMA, is proposed to accurately model the underlying structure of the input space through its transformation into feature representations in three distinct manifolds.
\item The proposed MMA can be added to any Vision Transformer to better guide its attention towards important appearance, color and texture features of an image.
\item Extensive experimentation with various ViTs and on different image classification and semantic segmentation datasets verifies the performance improvements achieved with the proposed MMA.
\end{itemize}

\section{Related work}
\label{sec:rwork}

\subsection{Vision Transformers}

Dosovitskiy \emph{et al.}, in \cite{Author3}, were the first to propose ViT as a pure Transformer backbone for image classification. In their work, the input image is segmented into a series of non-overlapping image patches that are then projected into a linear embedding sequence. This sequence is concatenated with a learnable positional encoding that holds information on the spatial location of patches in the image before being fed to the Transformer for classification.

Since then, ViTs have been heavily modified to achieve improved performance on computer vision tasks \cite{Author16}. Several ViTs combine CNNs with Transformers to effectively leverage the convolutional biases in images and enhance accuracy. Based on empirical observations that CNNs are better teachers than Transformers, Touvron \emph{et al.} employed a teacher-student strategy to transfer the inductive bias of the CNN teacher to the Transformer student through knowledge distillation \cite{Author4}. The authors in \cite{Author5} proposed ConViT, in which a parallel convolution branch was attached to the Transformer branch to impose convolutional inductive biases via a Gated Positional Self-Attention that approximates the locality of convolutional layers. Yuan \emph{et al.} proposed the Convolution-enhanced Image Transformer that uses a convolutional module to extract patch embeddings from low-level features and a layer-wise attention to model long-range dependencies \cite{Author6}.

To improve local attention and enhance the feature extraction capabilities of ViTs, Han \emph{et al.} combined Transformer networks at patch and pixel levels to produce better representations with rich local and global information \cite{Author7}. Meanwhile, the CSWin Transformer \cite{Author50} achieves strong modeling capabilities by performing self-attention in horizontal and vertical stripes in parallel, while A-ViT \cite{Author49} adaptively adjusts the inference cost of ViTs for images of different complexity by reducing the number of processed tokens as inference proceeds. In a similar fashion, the authors in \cite{Author66} combined locally-grouped self-attention and global sub-sampled attention in a Vision Transformer, achieving competitive image classification and object detection results. To further reduce computational cost, several works approximated the quadratic operations in self-attention by low-rank matrices \cite{Author33}, positive orthogonal random features \cite{Author35}, locality-sensitive hashing \cite{Author34} and memory-efficient coupling attention maps \cite{Author36}. On the other hand, the authors in \cite{Author63} replaced the attention mechanism with a spatial pooling operation, demonstrating that the resulted MetaFormer can achieve state-of-the-art performance in several computer vision tasks.

Realizing that ViTs need sufficiently large datasets to perform well, Hassani \emph{et al.} proposed three compact ViT architectures, namely ViT-Lite, Compact Vision Transformers (CVTs) and Compact Convolutional Transformers (CCTs) \cite{Author1}. ViT-Lite is similar to the original ViT but with smaller patch sizing suitable for small datasets. CVTs expand on ViT-Lite by pooling the sequential information from the Transformer, eliminating the need for the extra classification token. Finally, CCTs further expand on CVTs by adding convolutional blocks to the tokenization step, thus preserving local information while encoding relationships between patches, unlike the original ViT.

Recognizing that a fixed resolution across the entire network neglects fine-grained information and brings heavy computational costs, several works proposed hierarchical structures for ViTs. Yuan \emph{et al.} proposed the hierarchical structuring of the patch embeddings through the combination of neighbouring embeddings into a single one \cite{Author8}. Similarly, the Token Pyramid Vision Transformer produces scale-aware semantic features by processing tokens from various scales into a single token to augment the representation \cite{Author51}. Other ViTs reduce the spatial dimensions of the output progressively, similarly to CNNs, using spatial-reduction attention or pooling layer \cite{Author10,Author11}. The Swin Transformer employs shifting windows in each layer to create hierarchical global and boundary features through cross-window interactions \cite{Author9, Author65}, while the authors in \cite{Author64} proposed CrossFormer that employs a cross-scale embedding layer and a long-short distance attention that builds dependencies among neighboring embeddings as well as embeddings that are far away from each other. Following a different strategy, the authors in \cite{Author56} proposed the ViTAE architecture that employs convolution blocks in parallel to the multi-head self-attention modules to enable the Transformer network to better learn local features and global dependencies collaboratively. Whereas, the authors in \cite{Author67} proposed the use of deformable attention to model the relations among tokens under the guidance of the important regions in the feature maps, achieving very accurate results on image classification and dense prediction tasks. Finally, UniFormer was proposed in \cite{Author57} to combine the advantages of 3D convolution and spatiotemporal self-attention, achieving a balance between computation efficiency and accuracy. Different from traditional transformers, UniFormer can tackle both spatiotemporal redundancy and dependency by learning local and global token affinity respectively in shallow and deep layers. 

Traditionally, ViTs process raw pixel intensities directly in the Euclidean space without considering how different data representations may affect their performance. The proposed work improves local attention through the use of feature representations in different manifolds to create more descriptive attention maps.

\subsection{Manifold Background}

A manifold is a topological space that locally resembles a Euclidean space near each point \cite{Author41}. Essentially, a manifold is a mapping from one space to another, allowing similar features to appear closer to each other, while dissimilar features move further apart. Manifolds have been widely employed in computer vision tasks due to their ability to model different and complementary statistical and geometrical properties of features that can be beneficial for a given task \cite{Author2,Author51,Author54}. Two widely employed special types of manifolds used to describe image sets and videos in the literature are the SPD and Grassmann manifolds.

Huang \emph{et al.} utilized properties of the Riemmanian geometry on manifolds and proposed a new similarity method based on SPD features for clustering tasks \cite{Author2}. Yu \emph{et al.} proposed the contour covariance that lies on the SPD manifold as a region descriptor for accurate image classification \cite{Author17}. Similarly, Chu \emph{et al.} proposed the modelling of image sets with covariance matrices for improved classification performance \cite{Author20}. The importance and usefulness of feature modelling on the SPD manifold can be further highlighted from the design of novel deep networks and network layers, such as Variational Autoencoders \cite{Author52}, LSTMs \cite{Author53}, GRUs \cite{Author19} and mapping and pooling layers \cite{Author43} to handle and learn from features on the SPD manifold. The main drawback of employing covariance features in a deep learning framework is the non-linearity of the SPD manifold that introduces challenging optimization problems \cite{Author53}.

On the other hand, Grassmannian geometry has been widely employed in several computer vision tasks, such as image classification and action recognition. Data representations that lie on the Grassmann manifold include the Vectors of Locally Aggregated Descriptors (VLAD) that have been successfully employed for medical image classification \cite{Author12} and the Linear Dynamic System (LDS) features that have been employed for fire and smoke detection in video sequences \cite{Author45} and skeletal sequence modelling for sign language and action recognition \cite{Author14,Author18}. In a different approach, Dimou \emph{et al.} introduced LDS features in a ResNet architecture, achieving state-of-the-art performance in image classification \cite{Author46}. Finally, there are works on the clustering of image sets modelled in high-dimensional Grassmann manifold through the projection to a low-dimensional space using an unsupervised dimensionality reduction algorithm based on Neighborhood Preserving Embeddings \cite{Author21} or a new low rank approximation model that relies on the Double Nuclear norm \cite{Author55}.

To further leverage the discriminative power of feature representations in different manifolds, Wang \emph{et al.} fused SPD and Grassmann manifold representations for clustering purposes, achieving state-of-the-art performance \cite{Author47}. However, current ViTs have largely overlooked alternative data representations, typically focusing only on the Euclidean space of pixel intensities. To address this issue, this work proposes MMA that combines feature representations in three distinctive manifolds to learn a highly descriptive attention map that can better identify the important context of input images. By leveraging the statistical properties of different manifolds, MMA can guide any Vision Transformer to better model the underlying structure of the input space, leading to improved classification and segmentation results.

\section{Proposed method}
\label{sec:method}

Inspired by the need for more descriptive attention maps, this work proposes MMA as a novel self-attention mechanism that leverages the statistical and geometrical properties of different manifolds to produce richer feature representations. In contrast to the vanilla self-attention mechanism, MMA projects the high-dimensional Euclidean space to the Grassmann and SPD manifolds and computes the respective distance maps on them, as shown in Fig. \ref{fig:motivation}. Then, a fused attention map is produced through the merging of the Euclidean, SPD and Grassmannian distance maps that is more attentive towards important appearance, color and texture features of the image, ultimately leading to more accurate results in different computer vision tasks.

\subsection{Preliminaries}

Given an input image $\textbf{I} \in \mathbb{R}^{H\times W \times C}$ with height $H$, width $W$ and $C$ channels, a Vision Transformer divides the image into non-overlapping rectangular patches of size $P \times P$ and then linearly project them into the space of the hidden dimension $D$ of the Transformer or employs a 2D convolution to perform both image patch extraction and linear projection. The result is a vector of patch embeddings $\textbf{X} \in \mathbb{R}^{L\times D}$, where $L = \frac{HW}{P^2}$ is the sequence length.

Position embeddings are then added to the patch embeddings to provide a sense of order in the input sequence, allowing ViT to model both the content of the patches and their spatial location with respect to other image patches. The most common position embeddings are either vectors with sine and cosine frequencies \cite{Author3} or learned embeddings \cite{Author5,Author32}.

The embeddings are finally fed to the self-attention mechanism that is the most important layer of a Transformer, responsible for computing an output representation that is more attentive towards important features of the input. Vaswani et al. proposed the multi-head self-attention (MHSA) that performs different linear projections of the input at different subspaces through parallel attention layers, called heads, concatenated together \cite{Author29}. MHSA is computed as follows:

\begin{gather}
    \textbf{Q}_i = \textbf{XW}_{q}^i, ~~ \textbf{K}_i = \textbf{XW}_{k}^i, ~~ \textbf{V}_i = \textbf{XW}_{v}^i\\
    \textbf{S}_i = Attention(\textbf{Q}_i, \textbf{K}_i, \textbf{V}_i),~~ i = 1, 2, \dots, h\\
    MHSA(\textbf{Q},\textbf{K},\textbf{V}) = concat(\textbf{S}_1, \textbf{S}_2, \dots, \textbf{S}_h)\textbf{W}_{o}
    \label{eq:mhsa}
\end{gather}

where $h$ is the number of heads, $\textbf{W}_o \in \mathbb{R}^{D\times D}$ is the weight projection matrix, $d=D/h$ is the feature dimension in each head, $\textbf{S}_i \in \mathbb{R}^{L \times L}$ is the attention map of each head and $\textbf{W}_{q}^i,~\textbf{W}_{k}^i,~\textbf{W}_{v}^i \in \mathbb{R}^{D\times d}$ are the weight matrices for the query, key and value vectors of each head $i$, respectively. The attention map of each head $\textbf{S}_i$ is equal to:

\begin{equation}
    \textbf{S}_i = \mathrm{softmax}~(\frac{\textbf{Q}_i \textbf{K}_i^T}{\sqrt{d}})\textbf{V}_i
    \label{eq:sdp}
\end{equation}

MHSA allows ViT to jointly attend to information from different representation subspaces at different positions, enabling it to gather more positional data because each head focuses on various regions of the input and creating a more comprehensive representation after the combination of the vectors.

\subsection{Multi-manifold Multi-head Attention (MMA)}

\begin{figure}[t]
  \centering
  \includegraphics[width=\linewidth]{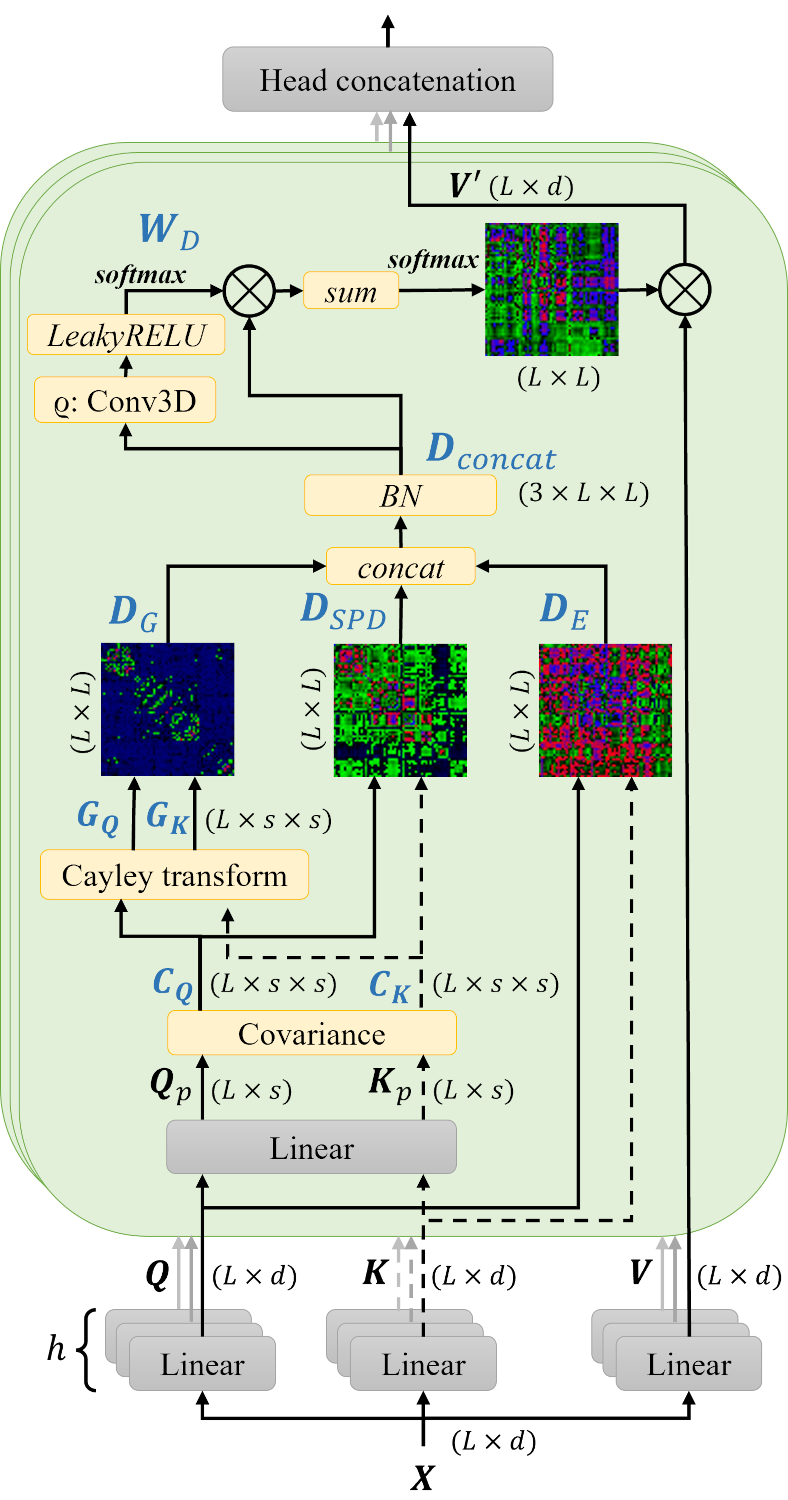}
  \caption{Proposed MMA architecture. The symbol $\otimes$ denotes matrix multiplication, while $BN$ is batch normalization.}
  \label{fig:mma}
\end{figure}

MMA can improve the performance of any ViT by replacing its vanilla self-attention mechanism. To achieve this, MMA transforms the input sequence to points in the Euclidean, SPD and Grassmann manifolds and computes distances between them in these manifolds in order to produce a fused attention map of high discriminative power, as shown in Fig. \ref{fig:mma}. Next, each manifold is described in detail, along with how the individual distance maps are computed and fused to form the refined attention map.

\subsubsection{Euclidean Manifold}

Given query $\textbf{Q} \in \mathbb{R}^{L \times d}$ and key $\textbf{K} \in \mathbb{R}^{L \times d}$ vectors, the Euclidean distance map of MMA is computed similarly to a vanilla Vision Transformer as:

\begin{equation}
    \textbf{D}_{E}(\textbf{Q},\textbf{K}) = \frac{\textbf{Q}\textbf{K}^T}{\sqrt{d_k}}
    \label{eq:emp}
\end{equation}

The distance map $\textbf{D}_{E} \in \mathbb{R}^{h\times L \times L}$, with $h$ representing the number of heads, expresses the similarity between query and key vectors, with higher values denoting greater distances between the two vectors in the Euclidean manifold.

\subsubsection{SPD Manifold}

The SPD manifold is a specific type of Riemann manifold composed of points expressed as square matrices $\textbf{M}$ of size $d\times d$ and it is denoted as:

\begin{equation}
    \mathcal{S}_{++}^d = \{ \textbf{M} \in \mathbb{R}^{d\times d}: \textbf{u}^T\textbf{M}\textbf{u} > 0 ~ 	\forall ~\textbf{u} \in \mathbb{R}^{d} - \{0_d\}\}
\end{equation}

For a matrix to be considered as point in a SPD manifold, it should be symmetrical and have positive eigenvalues. Covariance matrices possess such properties and thus they can be considered points in a SPD manifold. Covariance matrices have been widely employed in the literature to model appearance and texture features in computer vision tasks \cite{Author17,Author42}. As a result, the inclusion of covariance matrices in the computation of MMA is beneficial to the performance of a Vision Transformer due to incorporating additional information about the input, enhancing the discriminative power of the output feature representation. Several metrics can be used to measure the distance between points in a SPD manifold~\cite{Author13}, however, this work employs the Frobenius distance as it is not restricted by the values of the elements in the covariance matrices, unlike log-based distances.

Given query $\textbf{Q} \in \mathbb{R}^{L \times d}$ and key $\textbf{K} \in \mathbb{R}^{L \times d}$ vectors, a learnable linear projection operation is initially employed to reduce the dimensionality of the vectors and improve the computational efficiency of the proposed approach. The projected query and key vectors can be defined as $\textbf{Q}_p, \textbf{K}_p \in \mathbb{R}^{L \times s}$ with $s$ being the projection dimension. Afterwards, the covariance matrices of these vectors are computed as:

\begin{align}
    \textbf{C}_Q &= cov(\textbf{Q}_p) = \textbf{E}[(\textbf{Q}_p - \textbf{E}[\textbf{Q}_p])(\textbf{Q}_p - \textbf{E}[\textbf{Q}_p])^T]\label{eq:cq}\\
    \textbf{C}_K &= cov(\textbf{K}_p) = \textbf{E}[(\textbf{K}_p - \textbf{E}[\textbf{K}_p])(\textbf{K}_p - \textbf{E}[\textbf{K}_p])^T]\label{eq:ck}
\end{align}

Due to their properties, each of the matrices $\textbf{C}_Q$, $\textbf{C}_K \in \mathbb{R}^{L\times s\times s}$ describes a cluster of $L$ covariance matrices that lie as points on the SPD manifold. The SPD distance between the $i$-th covariance matrix of query and the $j$-th covariance matrix of key is then calculated as:

\begin{equation}
    \textbf{D}_{SPD}^{ij}(\textbf{C}_Q^i,\textbf{C}_K^j) =  \frac{{|| \textbf{C}_Q^i -\textbf{C}_K^j ||}_F}{s}
    \label{eq:spddm}
\end{equation}

where ${|| \cdot ||}_F$ denotes the Frobenius norm. Similar to the Euclidean distance map, $\textbf{D}_{SPD} \in \mathbb{R}^{h\times L \times L}$ expresses the similarity between query and key vectors and quantifies the distances between the two vectors in the SPD manifold.

\subsubsection{Grassmann Manifold}

The Grassmann manifold is another well-known special type of Riemann manifold that embeds all $p$-dimensional linear subspaces that lie in a $d$-dimensional Euclidean space. The Grassmann manifold, denoted as $ \mathcal{G}(p,d)$, can be represented by the set of orthogonal matrices from the orthogonal group $\mathcal{O}(p)$ as follows:

\begin{equation}
    \mathcal{G}(p,d) = \{ \textbf{X}\in \mathbb{R}^{d\times p} : \textbf{X}^T\textbf{X}=\textbf{I}_p \}/ \mathcal{O}(p),
\end{equation}

where \textbf{X} represents any point on the Grassmann manifold. Grassmann manifolds have been commonly employed to model sequential and time-varying signals as any linear dynamic system can be easily transformed to a point in the Grassmann manifold \cite{Author12,Author14}. As a result, the transformation of the input space to points in the Grassmann manifold can provide to a Vision Transformer additional information regarding texture and color variations in an image patch, leading to enriched feature representations with high discriminative power.

Several metrics have been defined to measure the distance between Grassmmanian points. The most common technique is to embed the manifold into the space of symmetric matrices with the mapping $\Pi : \mathcal{G}(p,d) \rightarrow \mathrm{Sym}(d), ~ \Pi(\textbf{X}) = \textbf{X}\textbf{X}^T$, which is a one-to-one, continuous and differentiable mapping \cite{Author31}. Moreover, to avoid the computation of the coordinates of all projected data and their pairwise distances as well as to improve efficiency, the kernel form of the projection distance \cite{Author15} is adopted.

Given query $\textbf{Q} \in \mathbb{R}^{L \times d}$ and key $\textbf{K} \in \mathbb{R}^{L \times d}$ vectors, the covariance matrices of \eqref{eq:cq} and \eqref{eq:ck} are first obtained. Since the orthogonality of the covariance matrices $\textbf{C}_Q$ and $\textbf{C}_K$ is not guaranteed, a transformation is required to ensure that the new matrices are orthogonal and thus represent points in the Grassmann manifold. One of the most popular technique to achieve orthogonality is through QR decomposition \cite{Author28}, however this work employs the Cayley transform \cite{Author58} due to its much smaller computational cost. According to this transform, the covariance matrices $\textbf{C}_Q$ and $\textbf{C}_K$ are initially transformed to the skew-symmetric matrices $\textbf{C}_Q^{skew}$ and $\textbf{C}_K^{skew}$ as follows:

\begin{align}
    \textbf{C}_Q^{skew} &= \frac{1}{2}(\textbf{C}_Q - \textbf{C}_Q^T) \\
    \textbf{C}_K^{skew} &= \frac{1}{2}(\textbf{C}_K - \textbf{C}_K^T)
\end{align}

Then, the Cayley map is employed to the skew-symmetric matrices to produce the following matrices:

\begin{align}
    \textbf{G}_Q &= (\textbf{I}_s + \frac{\textbf{C}_Q^{skew}}{2})(\textbf{I}_s - \frac{\textbf{C}_Q^{skew}}{2})^{-1} \\
    \textbf{G}_K &= (\textbf{I}_s + \frac{\textbf{C}_K^{skew}}{2})(\textbf{I}_s - \frac{\textbf{C}_K^{skew}}{2})^{-1}
\end{align}

The matrices $\textbf{G}_Q, \textbf{G}_K \in \mathbb{R}^{L \times s \times s}$ correspond to the representation of the query and key vectors in the Grassmann manifold and describe clusters of $L$ orthogonal matrices that lie on the manifold. Finally, the projection distance is employed to calculate the scaled distance between the $i$-th orthogonal matrix of query and the $j$-th orthogonal matrix of key in the Grassmann manifold as follows:

\begin{equation}
    \textbf{D}_{G}^{ij}(\textbf{G}_Q^i,\textbf{G}_K^j) = \frac{{||\textbf{G}_Q^i(\textbf{G}_Q^{i})^T - \textbf{G}_K^j(\textbf{G}_K^j)^T||}_F}{s}
    \label{eq:gdm}
\end{equation}

where ${|| \cdot ||}_F$ denotes the Frobenius norm. As with the other manifold distance maps, $\textbf{D}_{G} \in \mathbb{R}^{h\times L \times L}$ expresses the similarity between query and key vectors and quantifies the distances between the two vectors in the Grassmann manifold.

\subsubsection{Fusion of Manifolds}

After the computation of the individual distance maps $\textbf{D}_{E}$, $\textbf{D}_{SPD}$ and $\textbf{D}_{G} \in \mathbb{R}^{h\times L \times L}$ in each manifold, two setups, denoted as early and late fusion, are proposed to derive the output feature representation.

\textbf{Early Fusion.} In this setup, the three distance maps are concatenated together and then a 3D convolution operation $\varrho$ is employed to learn an element-wise weight matrix to perform an effective mapping of the distances in the different manifolds and generate the output feature representation. More specifically, the three distance maps are concatenated together forming the map:

\begin{equation}
\label{eq:map_concat}
   \textbf{D}_{concat} =  BN(concat(\textbf{D}_{E}, \textbf{D}_{SPD}, \textbf{D}_{G})) \in \mathbb{R}^{3 \times L \times L}
\end{equation}

with $BN$ denoting batch normalization. Afterwards, a weight matrix is computed to merge the different distance maps in an optimal way:

\begin{equation}
\label{eq:weight}
   \textbf{W}_D =  softmax(LeakyRELU(\varrho(\textbf{D}_{concat}))) \in \mathbb{R}^{3 \times L \times L}
\end{equation}

The weight matrix $\textbf{W}_D$ is responsible for computing appropriate weights in an element-wise manner to accurately merge the different distances. Finally, the output feature representation $\textbf{V}'$ is computed as follows:

\begin{equation}
\label{eq:early_fusion}
   \textbf{V}' =  softmax(sum(\textbf{W}_D \otimes \textbf{D}_{concat}))\textbf{V} \in \mathbb{R}^{L \times d}
\end{equation}

where the $sum$ operation is applied to the first dimension of the product of the weight matrix and the concatenated distance maps, effectively removing the first dimension.

\textbf{Late Fusion.} In this setup, three parallel attention mechanisms are employed, each processing the input in a different manifold, thus computing the manifold feature representations $\textbf{V}'_E=softmax(\textbf{D}_E)\textbf{V}$, $\textbf{V}'_{SPD}=softmax(\textbf{D}_{SPD})\textbf{V}$ and $\textbf{V}'_G=softmax(\textbf{D}_G)\textbf{V}$ for the Euclidean, SPD and Grassmann manifolds, respectively. Then, the output feature representation $\textbf{V}' \in \mathbb{R}^{L\times d}$ is equal to:

\begin{equation}
\label{eq:late_fusion}
    \textbf{V}' =  L(concat(\textbf{V}'_E, \textbf{V}'_{SPD}, \textbf{V}'_G))
\end{equation}

where $L$ performs a linear projection from the size of the concatenated feature representation $3d$ to the final size of the feature representation $d$.

\section{Experimental Results}
\label{sec:results}

\subsection{Implementation details}

MMA was introduced in several state-of-the-art hierarchical and compact ViTs to replace their vanilla self-attention mechanism. More specifically, the ViT-Lite-6/4, CVT-6/4 and CCT-7/3x2 models proposed in \cite{Author1}, the Swin-T model proposed in \cite{Author9} and the UniFormer-S model proposed in \cite{Author57} were employed, giving rise to new models with the same name and the MMA-* prefix, when MMA is utilized. For the Swin-T model, a patch size of $2$, a window size of $4$, an embedding dimension of $96$, an mlp ratio of $2$, depths of $(2,6,4)$ and number of heads equal to $(3,6,12)$ for the different layers were selected so that the model can be trained on the small datasets that was tested on. 

Experiments were conducted on 4 well-known image classification datasets, namely C-10 \cite{Author38}, C-100 \cite{Author38}, T-ImageNet \cite{Author40} and ImageNet, as well as on a popular semantic segmentation dataset, namely ADE20K \cite{Author59, Author60}. C-10 consists of $50$K training and $10$K test images equally distributed among $10$ classes, C-100 consists of $50$K training and $10$K test images equally distributed among $100$ classes, T-ImageNet consists of $100$K training and $10$K validation images equally distributed among $200$ classes, while ImageNet consists of more than $1.2$M training and $50$K validation images distributed among $1000$ classes. In T-ImageNet and ImageNet, the validation images were used to test the performance of ViTs. Finally, the Ade20K dataset consists of around $20$K training and $2$K test images depicting $150$ classes of objects.

\begin{table*}[t]
\centering
\caption{Ablation study using the MMA-ViT-Lite-6/4 model with the early and late fusion of manifolds}
\label{tab:ablation}
\setlength{\tabcolsep}{22pt}
\begin{tabular}{ccccccc}
\hline
\multicolumn{3}{c}{\textbf{Manifolds}} & \multicolumn{2}{c}{} & \multicolumn{2}{c}{\textbf{Accuracy (\%)}}\\
\textbf{E} & \textbf{SPD} & \textbf{G} & \textbf{Params (M)} & \textbf{FLOPs (G)} & \textbf{C-10} & \textbf{C-100} \\
\hline
\multicolumn{7}{c}{\textit{Early Fusion}} \\
\hline
\checkmark &   &   &   3.195   & 0.219   & 90.94    & 69.2  \\
& \checkmark  &    & 3.197   & 0.214   & 90.49    & 70.38 \\
&    & \checkmark  & 3.197   & 0.214   & 88.96    & 67.48 \\
\checkmark & \checkmark  &  & 3.197 & 0.221 & 91.89 & 72.41 \\
\checkmark &  & \checkmark & 3.197   & 0.221   & \textbf{92.85}    & 72.48 \\
& \checkmark  & \checkmark  & 3.197   & 0.215   & 92.03    & 71.35 \\
\checkmark & \checkmark & \checkmark  & 3.197 & 0.222 & 92.47 &\textbf{72.55} \\
\hline
\multicolumn{7}{c}{\textit{Late Fusion}} \\
\hline
\checkmark &   & \checkmark & 3.59 & 0.252 & 91.43 & 71.71 \\
\checkmark & \checkmark &   & 3.59 & 0.253 & 91.15 & 71.21 \\
& \checkmark & \checkmark & 3.59 & 0.246 & 89.3 & 68.25\\
\checkmark & \checkmark & \checkmark & 3.983 & 0.285 & 90.95 & 71.78 \\
\hline
\end{tabular}
\end{table*}
 
All experiments were run with a fixed batch size of $128$ and for $200$ epochs for C-10 and C-100 and $300$ epochs for T-ImageNet and ImageNet. For fair comparison, the implementation of \cite{Author1} was used for the experiments on C-10, C-100 and T-ImageNet and the implementation of \cite{Author57} was used for the experiments on ImageNet and ADE20K. As a result, the AdamW optimizer \cite{Author23} was used with a weight decay of $0.01$, a base learning rate of $5e^{-4}$ and a cosine learning rate scheduler that adjusts the learning rate during training \cite{Author24}. A warmup of 10 epochs was applied by increasing gradually the learning rate from 0 to the initial value of the cosine learning rate scheduler \cite{Author25}. Label smoothing \cite{Author22} with a probability $\epsilon=0.1$ was applied during training, where the true label is considered to have a probability of $1-\epsilon$ and the probability $\epsilon$ is shared between the other classes. For data augmentation, Auto-Augment \cite{Author26} was adopted to transform the training data with adaptive learnable transformations, such as shift, rotation, and color jittering. Moreover, the Mixup strategy \cite{Author27} was used to generate weighted combinations of random sample pairs from the training images.

Finally, the projection size for the covariance matrices $s$ was set to 4, the 3D convolution layer for the computation of the weight matrix had a kernel of size $1\times1\times1$ and the LeakyRELU function had a negative slope of 0.1. The code was implemented in PyTorch and all tested transformers with and without the proposed MMA were trained from scratch on a PC with 2 Nvidia 3090Ti gpus for fair comparison.

\subsection{Ablation study}
\label{sec:ablation}

\begin{figure}[t]
  \centering
  \includegraphics[width=\linewidth]{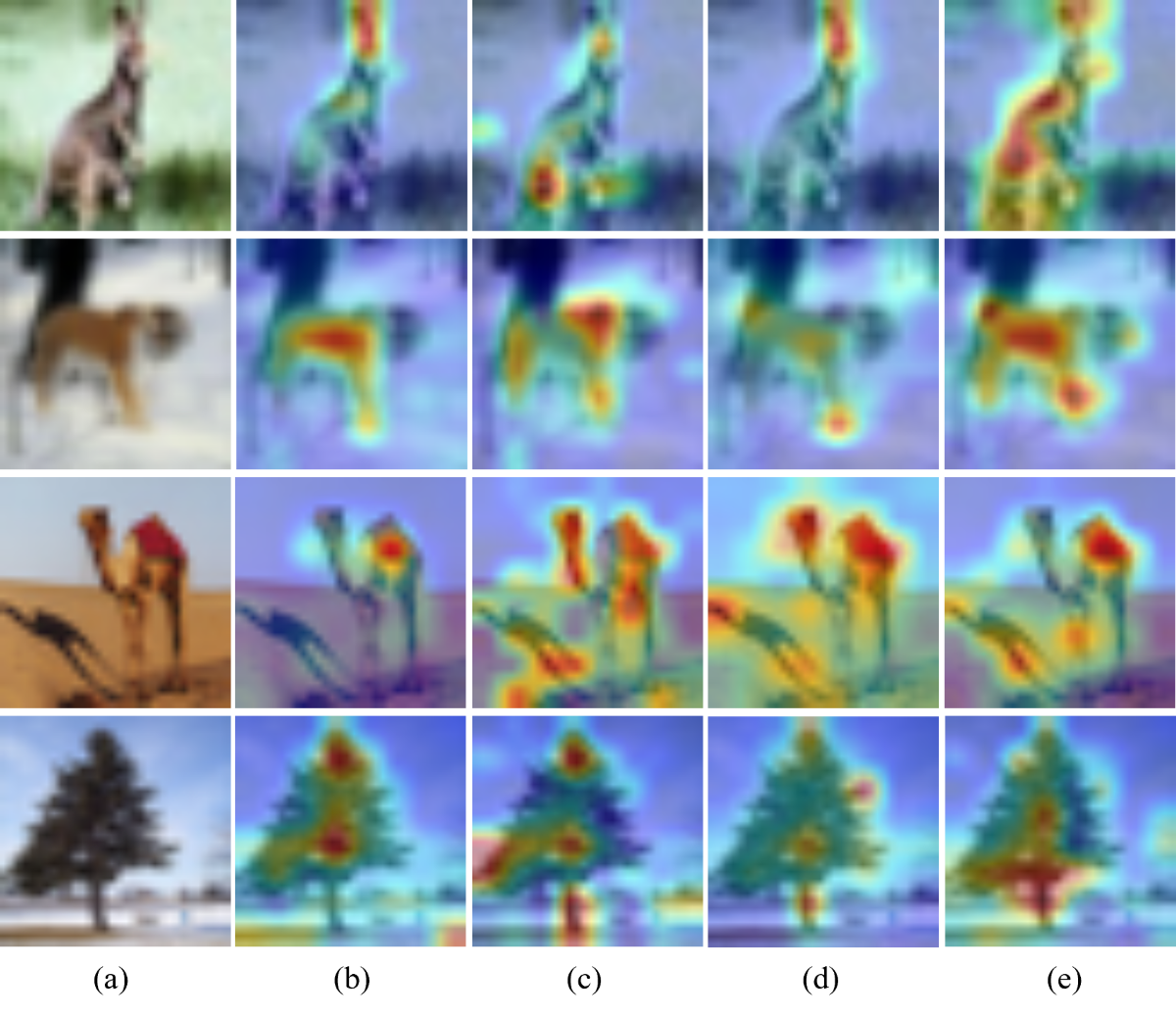}
  \caption{Class activations in C-100 using MMA-ViT-Lite-6/4. (a) Original images and class activation maps using (b) Euclidean, (c) SPD, (d) Grassmann and (e) all three manifolds.}
  \label{fig:c100_vis}
\end{figure}

To evaluate the contribution of each manifold to the performance of a Vision Transformer, experiments were conducted using the MMA-ViT-Lite-6/4 model in the C-10 and C-100 datasets. In these experiments, all possible combinations of manifold distance maps in early or late fusion setups were utilized and the results are presented in Table \ref{tab:ablation}.

\begin{table*}[t]
\centering
\caption{Comparative evaluation on C-10, C-100 and T-ImageNet. Parameters and FLOPs were computed for images of size $32\times32$.}
\label{tab:image_small}
\setlength{\tabcolsep}{17pt}
\begin{tabular}{lccccc}
\hline
& & & \multicolumn{3}{c}{\textbf{Accuracy (\%)}}\\
\textbf{Method}  & \textbf{Params (M)} & \textbf{FLOPs (G)} & \textbf{C-10} & \textbf{C-100}  & \textbf{T-ImageNet} \\ 
\hline
ResNet-100 \cite{Author37} &1.70 & 0.25 & 93.39 & 72.78 & - \\
ResNet-164   \cite{Author37}    & 1.73  &    0.26   & 94.54  & 75.67 & -\\
EfficientNet-B0 \cite{Author39} & 3.70 & 0.12 & 94.66 & 76.04 & -  \\ \hline
Linformer \cite{Author33} & 3.96 & 0.28 & 92.45 & 70.87 & - \\
Performer \cite{Author35}& 3.85 & 0.28 & 91.58 & 73.11 & -\\
Reformer \cite{Author34} & 3.39 & 0.25 & 90.58 & 73.02 & - \\
Couplformer-7 \cite{Author36}& 3.85 & 0.28 & 93.44 &74.53 & -\\ \hline

ViT-Lite-6/4 \cite{Author1}     & 3.195   &    0.219   & 90.94  & 69.2 & 49.18\\
\textbf{MMA-ViT-Lite-6/4} &     3.197 &  0.222   & 92.47   &   72.55 & 53.16 \\ \hline
CVT-6/4 \cite{Author1}          &  3.195 &   0.216  & 92.58  & 72.25   & 51.45\\
\textbf{MMA-CVT-6/4}      &  3.197  &        0.218   & 93.53  &     75.92  & 55.87\\ 
\hline
Swin-T \cite{Author9}          & 7.049 & 0.243 & 91.88  & 72.34   & 60.64 \\
\textbf{MMA-Swin-T}      & 7.05 & 0.246 &  92.94 &  73.7  & 61.57 \\ 
\hline
CCT-7/3×2 \cite{Author1}        & 3.859 &  0.29  & 93.65  & 74.77  & 61.07 \\
\textbf{MMA-CCT-7/3×2}   &  3.861  &   0.293     & \textbf{94.74} & \textbf{77.5} & \textbf{64.41}\\ 
\hline
\end{tabular}
\end{table*}

From the results of Table \ref{tab:ablation}, it can be concluded that when the manifolds are utilized on their own, the Euclidean and SPD manifolds perform similarly and slightly better than the Grassmann manifold. However, when two or more manifolds are combined, the performance of MMA-ViT-Lite-6/4 is significantly improved in both C-10 and C-100 datasets. The best results are achieved when all three manifolds are employed, as it is seen from the performance of MMA-ViT-Lite-6/4 in the more challenging C-100 dataset. Noteworthy is the fact that the increase in the performance of MMA-ViT-Lite-6/4 is simultaneously accompanied by a negligible increase in the number of network parameters and a small increase in the number of floating point operations (FLOPs). Similar observations can be made when the late fusion of the manifold representations is performed. These results verify that the SPD and Grassmann manifolds contain different and supplementary information to the Euclidean manifold through the modelling of the appearance, color and texture variations in images. This information can guide a Vision Transformer to become more attentive towards a better modelling of the underlying input space, enabling it to achieve improved performance.

On the other hand, a comparison between early and late fusion of manifolds reveals that early fusion leads to superior performance in both datasets, while utilizing almost $25\%$ fewer network parameters and around $28\%$ fewer FLOPs. These results show that fusing the manifold distance maps inside the Transformer ensures the generation of a refined attention map with high discriminative power.

To further highlight the benefits of employing additional manifolds, a visualization of class activation maps in a few images from the C-100 dataset is illustrated in Fig. \ref{fig:c100_vis}. It can be observed that although the vanilla self-attention allows the model to pay attention on the ears of a kangaroo or the hump of a camel, the proposed MMA enables the model to become highly attentive towards additional information, such as the entire body of the kangaroo and both the hump and the legs of a camel, thus increasing the confidence of the model in its classification results. For the rest of the experiments, it is assumed that the proposed MMA employs the early fusion of all three manifolds since this combination leads to the optimal performance.

\subsection{Image classification on small datasets}

\begin{figure*}[t]
  \centering
  \includegraphics[width=\linewidth]{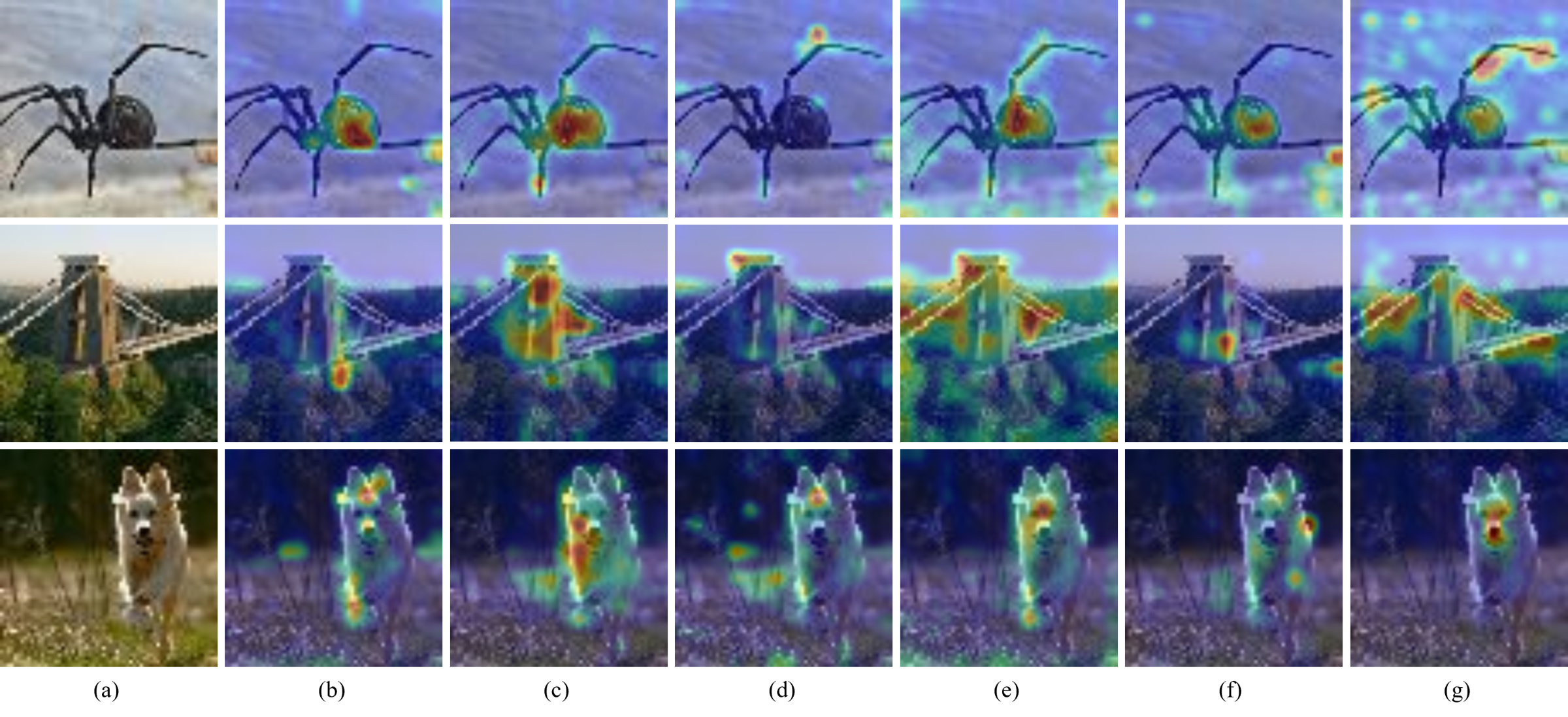}
  \caption{Class activations in T-ImageNet. (a) Original images and class activation maps using (b) CVT-6/4, (c) MMA-CVT-6/4, (d) Swin-T, (e) MMA-Swin-T, (f) CCT-7-3x2 and (g) MMA-CCT-7-3x2 models.}
  \label{fig:tim_vis}
\end{figure*}

\begin{figure}[t]
  \centering
  \includegraphics[width=\linewidth]{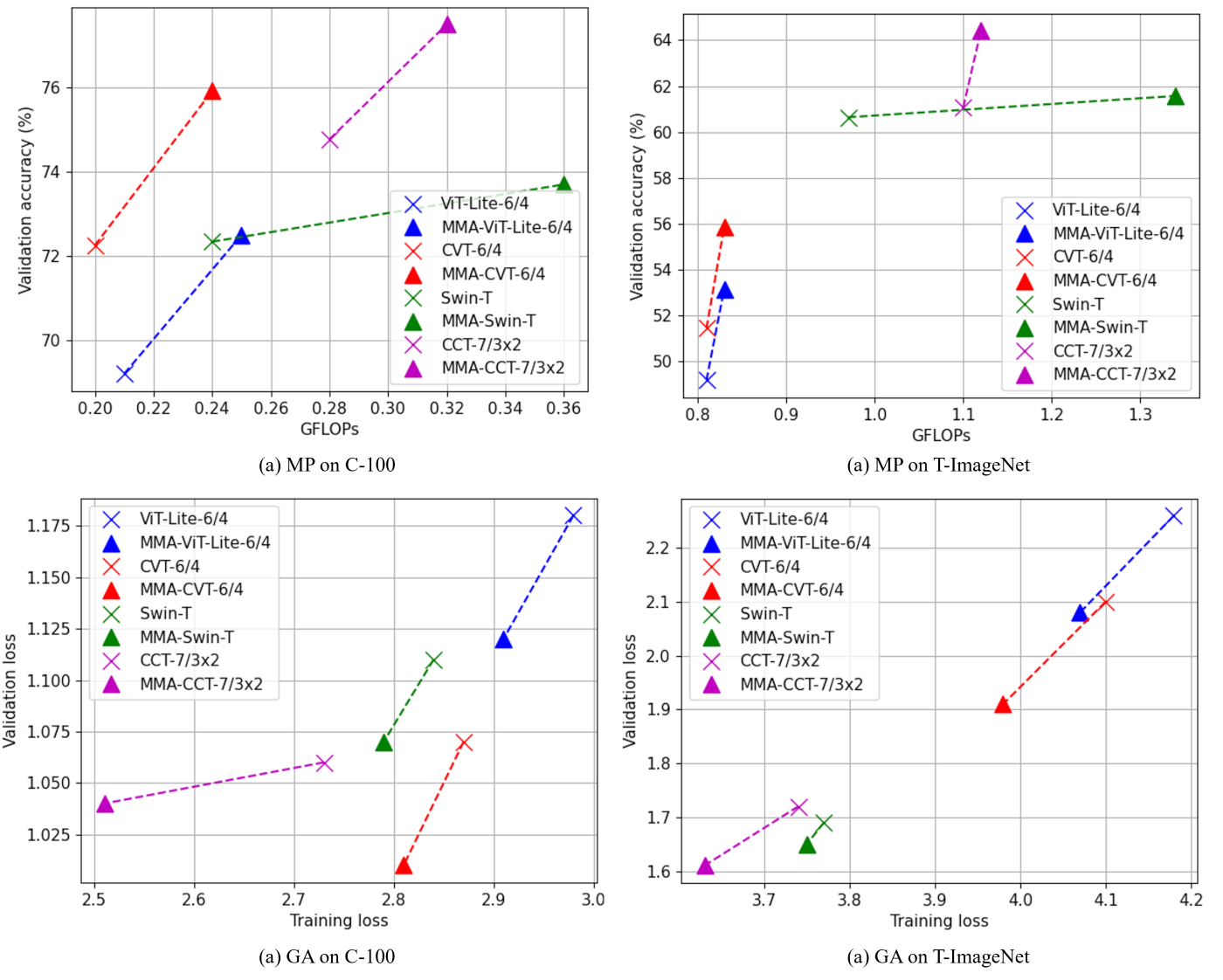}
  \caption{Impact of MMA on tested ViT variants trained on C-100 and T-ImageNet in terms of model performance (MP) and generalization ability (GA).}
  \label{fig:performance}
\end{figure}

\begin{figure}[t]
  \centering
  \includegraphics[width=\linewidth]{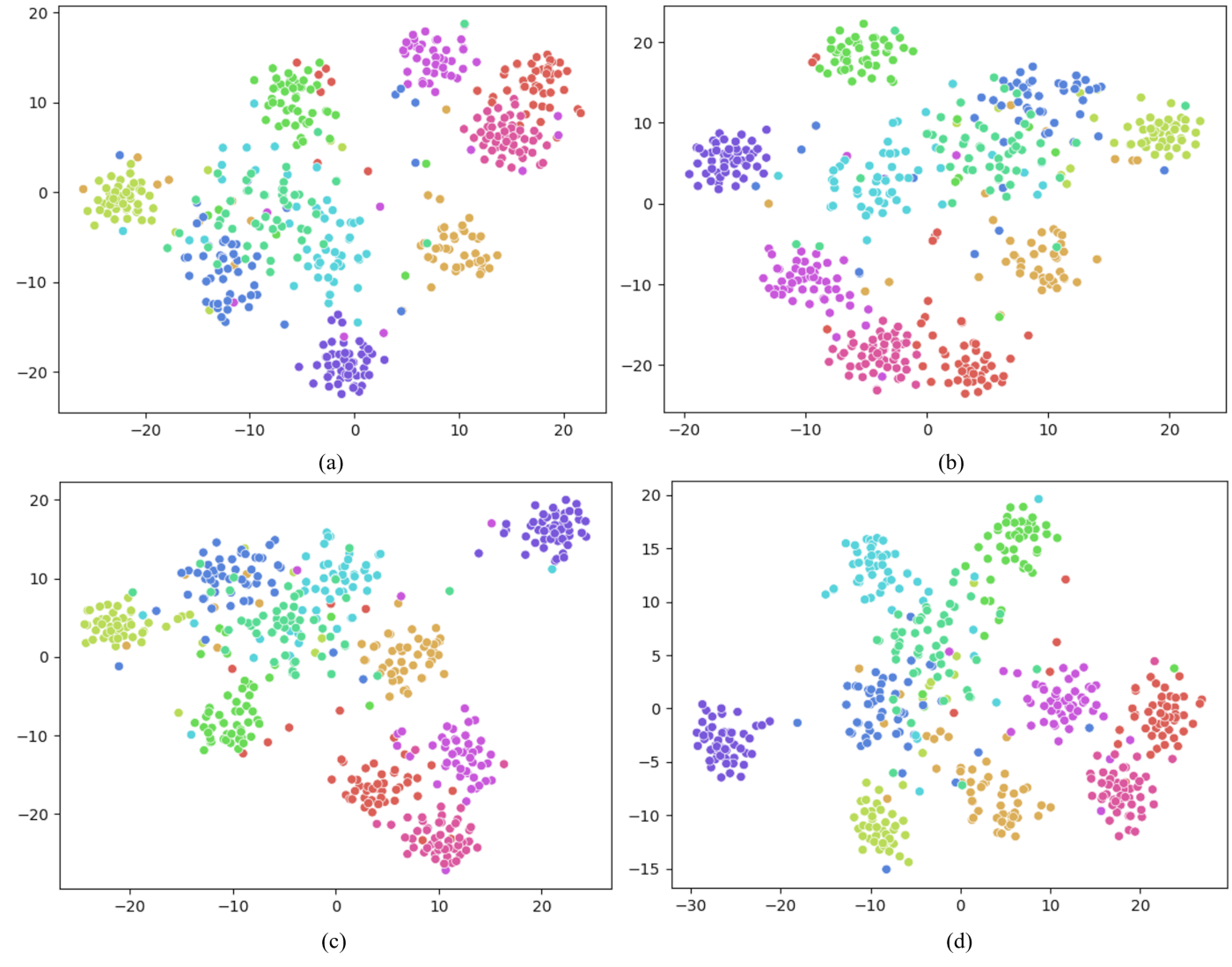}
  \caption{Visualization using t-SNE of outputs from (a) Swin-T, (b) MMA-Swin-T, (c) CCT-7-3x2 and (d) MMA-CCT-7-3x2 of 10 random classes of T-ImageNet.}
  \label{fig:tim_tsne}
\end{figure}

To demonstrate the benefits of MMA, a comparative evaluation of various hierarchical and compact ViTs with their vanilla self-attention and the proposed MMA in C-10, C-100 and T-ImageNet is performed. In addition, the selected Transformers are compared against other CNN networks and ViTs and the results are presented in Table \ref{tab:image_small}. The results demonstrate that the MMA-enhanced ViTs achieve superior performance in all datasets with respect to their original versions at the expense of a small increase in parameters and FLOPs. A comparison with other state-of-the-art deep networks shows that MMA-CCT-7/3x2 achieves the highest performance in all datasets, outperforming other ViTs and even well-established CNN networks, such as ResNet-164 \cite{Author37} and EfficientNet-B0 \cite{Author39}.

Furthermore, Fig. \ref{fig:tim_vis} depicts the class activation maps for CVT-6/4, Swin-T and CCT-7/3x2, and their MMA-* counterparts. It can be deduced that the MMA-enhanced ViTs become more attentive towards significant parts of the object of interest, such as the legs of a spider, the arch rib of a bridge and the eyes and noise of a dog, thus leading to more accurate classification results. Additional conclusions can be drawn from Fig. \ref{fig:performance} that presents the impact of MMA on the tested ViTs in terms of model performance and generalization ability. Irrespective of their network architecture, all ViTs are significantly improved with MMA through a decrease in the training and validation losses and an increase in the validation accuracy.

Finally, Fig. \ref{fig:tim_tsne} depicts the distribution of $10$ random classes of T-ImageNet using the t-distributed stochastic neighbor embedding (t-SNE) algorithm \cite{Author44} that is suitable for visualizing high-dimensional data by applying nonlinear dimensionality reduction. To this end, the feature vectors after the head concatenation of \eqref{eq:mhsa} are employed as the high-dimensional input to the t-SNE algorithm, which then computes a two-dimensional output. From the distribution of the $10$ T-ImageNet classes in the 2D space, it can be observed that MMA leads to more compact classes (i.e., points of the same class closer to each other) and less stray points for all tested ViTs. All these results verify the ability of MMA to guide any Vision Transformer to accurately model the underlying input space and produce highly descriptive output representations. This is achieved through the fusion of different and complementary manifold representations that enables a Vision Transformer to model the important context of an image through a refined attention map of high discriminative power. As a result, the proposed MMA can effectively substitute the vanilla self-attention in Vision Transformers, significantly improving their classification performance.

\subsection{Image classification on ImageNet}

\begin{figure}[t]
  \centering
  \includegraphics[width=\linewidth]{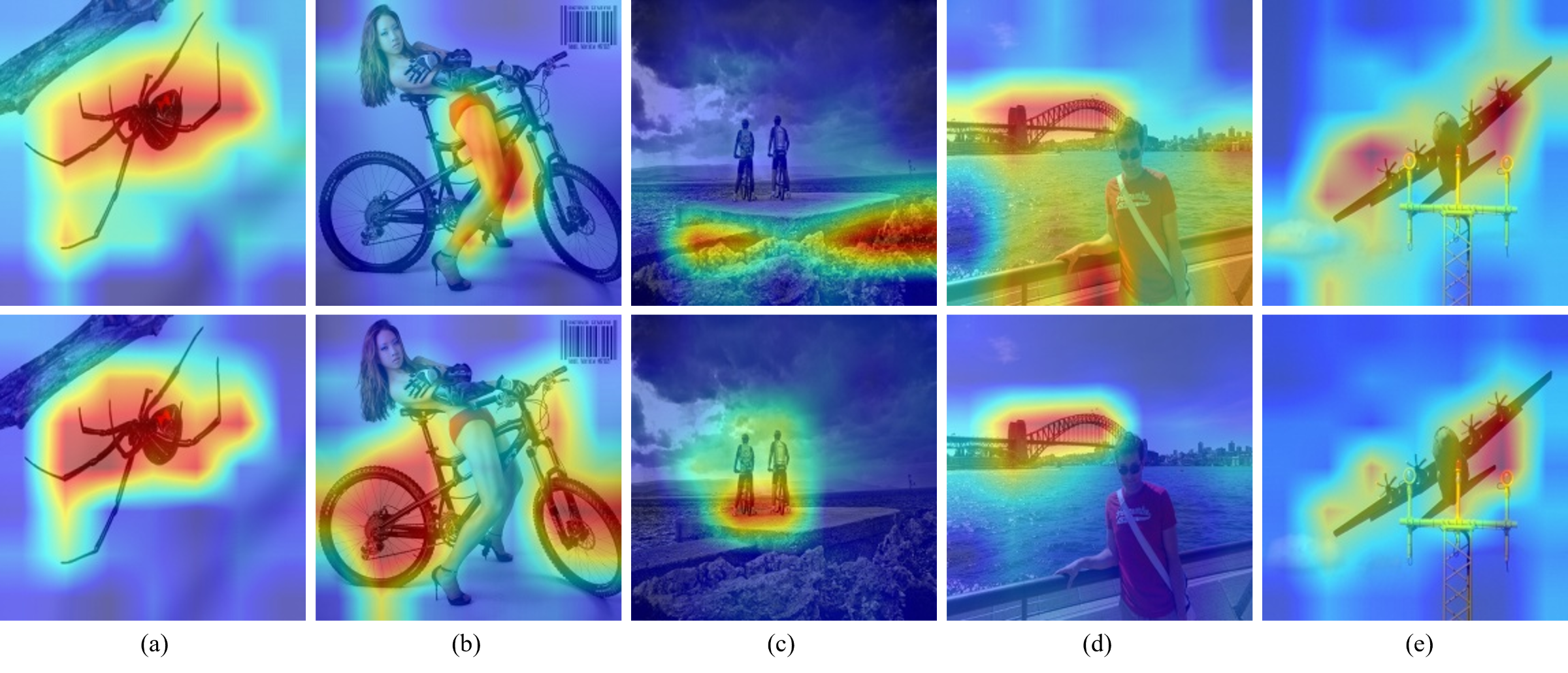}
  \caption{Activation maps from UniFormer-S (first row) and MMA-UniFormer-S (second row) for the classes (a) black widow spider, (b-c) mountain bike, (d) steel arch bridge and (e) warplane.}
  \label{fig:imnet_vis}
\end{figure}

\begin{table*}[t]
\centering
\caption{Comparative evaluation on ImageNet with images of size $224\times224$.}
\label{tab:image_imnet}
\setlength{\tabcolsep}{24pt}
\begin{tabular}{lcccc}
\hline
\textbf{Method}  & \textbf{Arch.} & \textbf{Params (M)} & \textbf{FLOPs (G)} & \textbf{Top-1 acc.}\\
\hline
ResNet-50 \cite{Author61} & & 25.6 & 3.8 & 76.7 \\
EfficientNet-B0 \cite{Author39} & & 5.3 & 0.4 & 77.1 \\
ResNet-101 \cite{Author61} & CNN & 44.5 & 7.6 & 78.3 \\
ResNet-152 \cite{Author61} & & 60.2 & 11.3 & 78.9 \\
ConvNeXt-T \cite{Author62} & & 28.0 & 4.5 & \textbf{82.1} \\
\hline
PVT-S \cite{Author10} & & 24.5 & 3.8 & 79.8\\
DeiT-S \cite{Author4} & & 22.1 & 4.6 & 79.9\\
Twins-PCPVT-S \cite{Author66} & & 24.1 & 3.8 & 81.2\\
Swin-T \cite{Author9} & & 29.0 & 4.5 & 81.3\\
ConViT-S \cite{Author5} & & 27.0 & 5.4 & 81.3\\
PoolFormer-S36 \cite{Author63} & & 30.8 & 5.0 & 81.4\\
T2T-ViT-14 \cite{Author8} & & 21.5 & 5.2 & 81.5\\
Crossformer-T \cite{Author64} & & 27.8 & 2.9 & 81.5\\
Twins-SVT-S \cite{Author66} & ViT & 24.0 & 2.9 & 81.7\\
SwinV2-T \cite{Author65} & & 28.0 & 5.9 & 81.8\\
DAT-T \cite{Author67} & & 28.3 & 4.6 & 82.0\\
CeiT-S \cite{Author6} & & 24.2 & 4.5 & 82.0\\
ViTAE-S \cite{Author56} & & 23.6 & 5.6 & 82.0\\
UniFormer-S \cite{Author57} & & 21.5 & 3.6 & 82.3\\
Crossformer-S \cite{Author64} & & 30.7 & 4.9 & 82.5\\
\textbf{MMA-UniFormer-S} & & 21.5 & 3.7 & \textbf{82.6}\\
\hline
\end{tabular}
\end{table*}

To demonstrate the ability of the proposed MMA to improve the performance of Vision Transformers even on large and more challenging image classification datasets, such as ImageNet, we employed one of the state-of-the-art ViTs, namely UniFormer \cite{Author57} and we replaced the attention mechanism of its small version (i.e., UniFormer-S) with the proposed MMA, thus obtaining the MMA-UniFormer-S version of the model. Table \ref{tab:image_imnet} presents the performance improvement achieved after the modification, as well as performs a comparison against other state-of-the-art small-sized Vision Transformers.

From the results, it can be seen that the MMA-UniFormer-S outperforms UniFormer-S by 0.3\% on ImageNet, when the proposed multi-manifold attention is introduced, managing to achieve a state-of-the-art performance on ImageNet with an accuracy of 82.6\%. The results verify the ability of the proposed MMA to improve the performance of Vision Transformers in larger and more challenging datasets. A visualization of the attention maps extracted from the original UniFormer-S and the modified MMA-UniFormer-S models on a few images of ImageNet is presented in Fig. \ref{fig:imnet_vis}. This figure further demonstrates that the use of additional manifolds in the attention mechanism of a Vision Transformer can lead to a more refined and descriptive attention map that can better describe the content of an image. Of notable importance are the cases of mountain bike identification (Fig. \ref{fig:imnet_vis}(b-c)), in which it can be seen that the MMA-UniFormer-S manages to correctly identify the object of interest, overcoming the mis-classification errors performed by the UniFormer-S.

\subsection{Semantic segmentation on ADE20K}

To further evaluate the ability of the proposed MMA mechanism to improve the performance of Vision Transformers on tasks different from image classification, we train and evaluate the MMA-UniFormer-S on semantic segmentation using the challenging scene parsing ADE20K dataset. More specifically, we employ the MMA-UniFormer-S, pre-trained on ImageNet, as a backbone to the semantic FPN network \cite{Author10} and we employ the training scheme of the UniFormer with 80K iterations for fair comparison among the two models. 

\begin{table*}[t]
\centering
\caption{Comparative evaluation on semantic segmentation using ADE20K.}
\label{tab:segmentation}
\setlength{\tabcolsep}{26pt}
\begin{tabular}{lcccc}
\hline
\textbf{Backbone} & \textbf{Head} & \textbf{Params (M)}  & \textbf{FLOPs (G)} & \textbf{mIoU}\\
\hline
ResNet50 \cite{Author61} & FPN & 29 & 183 & 36.7\\
PVT-S \cite{Author10} & FPN & 28 & 161 & 39.8\\
Swin-T \cite{Author9} & FPN & 32 & 182 & 41.5\\
DAT-T \cite{Author67} & FPN & 32 & 198 & 42.6\\
Twins-S \cite{Author66} & FPN & 28 & 144 & 43.2\\
Twins-PCPVT-S \cite{Author66} & FPN & 28 & 162 & 44.3\\
UniFormer-S \cite{Author57} & FPN & 25 & 247 & 46.0\\
DAT-S \cite{Author67} & FPN& 53 & 320 & 46.1\\
CrossFormer-S \cite{Author64} & FPN & 34 & 210 & 46.4\\
\textbf{MMA-UniFormer-S} & FPN & 25 & 259 & \textbf{46.9}\\
\hline
\end{tabular}
\end{table*}

\begin{figure*}[t]
  \centering
  \includegraphics[width=\linewidth]{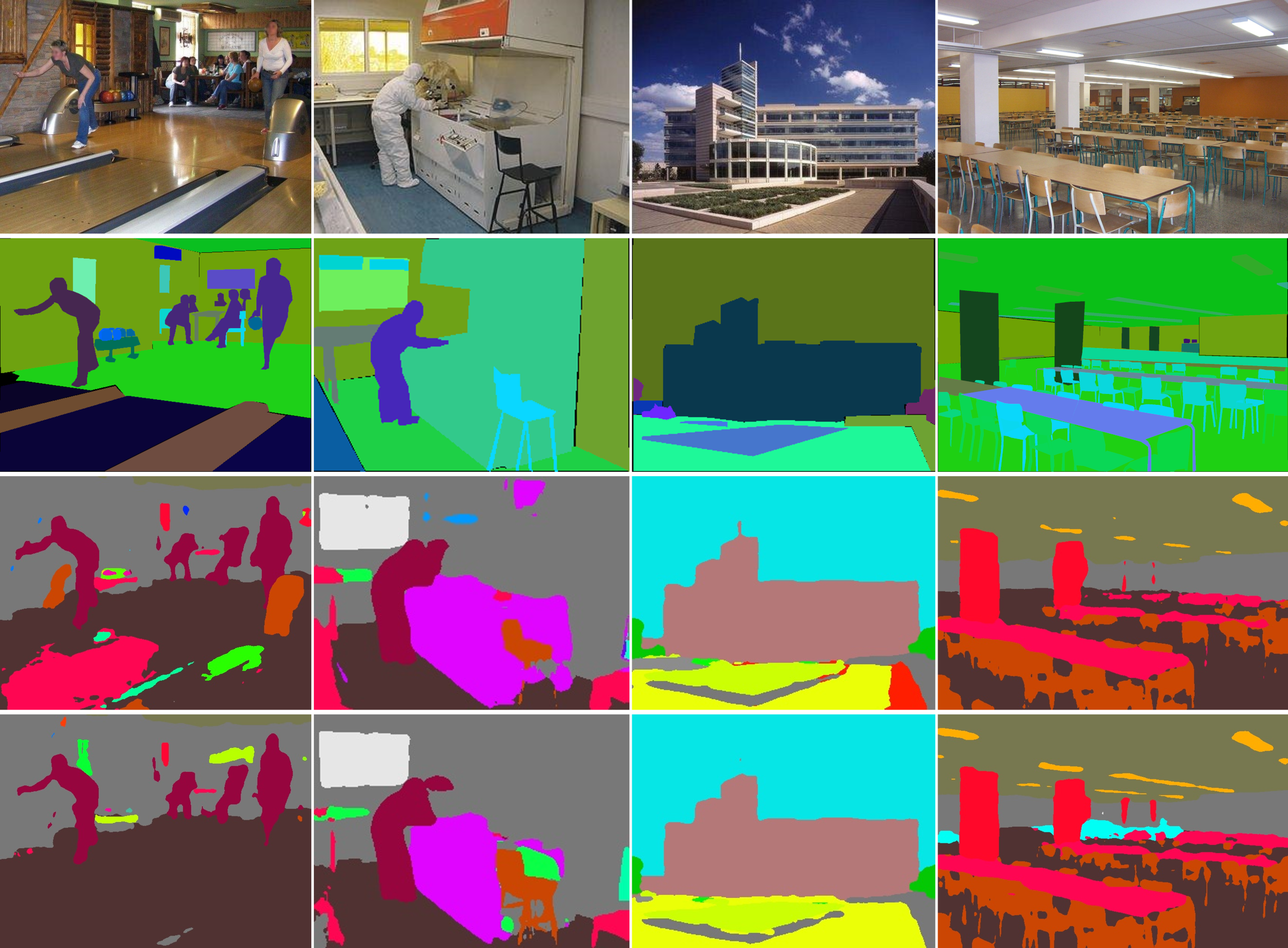}
  \caption{Semantic segmentation results in ADE20K. The first row depicts the original images, the second row shows the ground truth segmentation, while the third and fourth rows depict the segmentation results from the UniFormer-S + FPN and the MMA-UniFormer-S + FPN networks, respectively.}
  \label{fig:ade20k_vis}
\end{figure*}

From Table \ref{tab:segmentation}, it can be seen that the introduction of MMA-UniFormer-S as a backbone to the semantic FPN network can lead to significant improvements in the performance of the model in semantic segmentation. More specifically, the use of MMA-UniFormer-S leads to an improvement of 0.9\% in mIoU with respect to UniFormer-S, thus leading the network to achieve state-of-the-art performance with 46.9 mIoU, while retaining the smallest number of network parameters (25M) among other backbones. Finally, Fig. \ref{fig:ade20k_vis} visualizes a few segmentation results from the ADE20K dataset. These results further verify the higher accuracy achieved when the proposed MMA is introduced to the backbone model of a semantic segmentation network with the result being more accurately delineated objects. As a result, the proposed MMA can lead to performance improvements in both image classification and semantic segmentation with a minimal increase in network parameters and a slight increase in FLOPs.

\section{Conclusion}
\label{sec:conclusion}
This work proposes MMA as a novel self-attention mechanism that is suitable for any Vision Transformer irrespective of its architecture. The motivation behind MMA lies in the use of three distinct manifolds, namely Euclidean, SPD and Grassmann to model the input and produce a fused attention map that can more accurately attend to the important context of the input. Experimental results with hierarchical and compact ViT variants on several image classification datasets (i.e., C-10, C-100, T-ImageNet and ImageNet) and a challenging semantic segmentation dataset (i.e., ADE20K) verify the effectiveness of MMA in producing highly descriptive output representations and improving the performance of Vision Transformers in both image classification and semantic segmentation.

{\small
\bibliographystyle{ieee_fullname}
\bibliography{egbib}
}

\end{document}